\documentclass{article}

\PassOptionsToPackage{numbers, compress}{natbib}
\usepackage[preprint]{neurips_2026}

\usepackage[utf8]{inputenc} 
\usepackage[T1]{fontenc}    
\usepackage{hyperref}       
\usepackage{url}            
\usepackage{booktabs}       
\usepackage{amsfonts}       
\usepackage{nicefrac}       
\usepackage{microtype}      
\usepackage{xcolor}         

\usepackage{graphicx}
\usepackage{amsmath}
\usepackage{cleveref}
\usepackage{makecell}
\usepackage{enumitem}
\usepackage{tcolorbox}
\tcbuselibrary{breakable}
\usepackage{caption}
\usepackage{hyperref}
\usepackage{xurl}

\definecolor{icmlblue}{RGB}{28, 72, 127}    
\definecolor{icmllight}{RGB}{248, 250, 252} 
\definecolor{icmlaccent}{RGB}{191, 208, 230} 
\definecolor{chatgray}{RGB}{242, 247, 255} 
\definecolor{chatborder}{RGB}{56, 84, 145}  
\newtcolorbox{promptbox}[1][]{
  breakable,     
  colback=icmllight,          
  colframe=chatborder,          
  coltitle=white,             
  fonttitle=\bfseries,      
  title={#1},               
  arc=3mm,                  
  boxrule=0.5pt,           
  left=2mm, right=2mm, top=2mm, bottom=2mm, 
  fontupper=\small\ttfamily 
}

\title{AceGRPO: Adaptive Curriculum Enhanced Group Relative Policy Optimization for Autonomous Machine Learning Engineering}

\author{%
    Yuzhu Cai\textsuperscript{\rm 1,2 $\dagger$},
    Zexi Liu\textsuperscript{\rm 2 $\dagger$}, 
    Xinyu Zhu\textsuperscript{\rm 2 $\dagger$},
    Cheng Wang\textsuperscript{\rm 2}, 
    Yanfeng Wang\textsuperscript{\rm 2},
    Siheng Chen\textsuperscript{\rm 2,$\ast$}\\
    \textsuperscript{\rm 1} School of Computer Science and Engineering, Beihang University \\
    \textsuperscript{\rm 2} School of Artificial Intelligence, Shanghai Jiao Tong University \\
}

\begin{document}

\begingroup
    \renewcommand{\thefootnote}{\fnsymbol{footnote}} 
    
    \footnotetext[2]{Equal contribution. Order randomized.}

    \footnotetext[1]{Corresponding author: \texttt{sihengc@sjtu.edu.cn}}

\endgroup

\maketitle

\vspace{-2em}
\begin{abstract}
  Autonomous Machine Learning Engineering (MLE) requires agents to perform sustained, iterative optimization over long horizons. While recent LLM-based agents show promise, current prompt-based agents for MLE suffer from behavioral stagnation due to frozen parameters. Although Reinforcement Learning (RL) offers a remedy, applying it to MLE is hindered by prohibitive execution latency and inefficient data selection. Recognizing these challenges, we propose \textbf{AceGRPO} with two core components: (1) \textbf{Evolving Data Buffer} that continuously repurposes execution traces into reusable training tasks, and (2) \textbf{Adaptive Sampling} guided by a Learnability Potential function, which dynamically prioritizes tasks at the agent's learning frontier to maximize learning efficiency. Leveraging AceGRPO, our trained \textbf{Ace-30B} model achieves a \textbf{100\% valid submission rate} on MLE-Bench-Lite, approaches the performance of proprietary frontier models, and outperforms larger open-source baselines (e.g., DeepSeek-V3.2), demonstrating robust capability for sustained iterative optimization.
\end{abstract}

\vspace{-1em}
\section{Introduction}
\label{sec:intro}

Large Language Models (LLMs) have transitioned from static question-answering systems into autonomous agents capable of complex reasoning, tool use, and multi-step problem solving~\cite{tang2025eigen,wang2024openhands,zhang2025bohrium+}. 
This transition has enabled LLM agents to tackle tasks that demand not  single-shot answers but sustained iterative optimization. 
Autonomous Machine Learning Engineering (Autonomous MLE)—including settings  such as Kaggle competitions~\cite{chan2024mle}—exemplifies this class of challenges. 
Unlike standard Software Engineering where success is often binary (e.g., passing unit tests)~\cite{jimenez2023swe,du2025swe}, Autonomous MLE constitutes an empirical science requiring continuous refinement: agents must search high-dimensional hypothesis spaces, propose changes to architectures or data pipelines, and interpret noisy experimental feedback to improve performance over time~\cite{mlmaster}. 
Progress in this domain necessitates more than coding proficiency; it demands \textit{self-evolution}—the ability to sustain improvement trajectories, escape local optima, and refine strategies based on accumulated experience.

Most recent progress on Autonomous MLE relies on prompt-based methods~\cite{yang2025r,li2025fm,nam2025mlestar} to structure agent exploration at test time. While these methods scale inference-time search, they leave the underlying LLM parameters fixed. 
Because the policy does not update, the agent cannot convert trial-and-error experience into improved decision rules, and the policy stagnates. 
As a result, agents repeatedly explore suboptimal patterns even after thousands of episodes, yielding persistent behavioral plateaus. 
A natural alternative is to shift from \textit{searching} for solutions to \textit{training} better solvers using reinforcement learning (RL)~\cite{lu2025scaling}. 
Such a transition is essential to empower LLMs to sustain self-evolution and master the dynamics of iterative optimization in long-horizon MLE tasks~\cite{zhu2026toward}.
Leveraging the objective and verifiable metrics provided by MLE tasks~\cite{qiang2025mle}, Reinforcement Learning (RL) serves as a powerful method to steer this optimization process~\cite{guo2025deepseek}. 


However, establishing RL training for MLE remains an open challenge.
Success in MLE necessitates sustained iterative refinement unlike one-shot coding tasks~\cite{huang2023mlagentbench}. 
Training for such capabilities is hindered by data that is scarce and unevenly distributed~\cite{rengarajan2022reinforcement}. 
The challenge is further compounded by the prohibitive latency of execution, where a single step may require minutes to hours, rendering end-to-end full-trajectory RL computationally intractable~\cite{liu2025ml, deng2025supervised}.
Consequently, practical approaches must adopt step-wise rollouts, optimizing single actions conditioned on intermediate states to ensure training efficiency. 
Nevertheless, step-wise RL in Autonomous MLE faces two challenges:
(1) \textit{Where to sample states}: prior approaches often construct a large state pool by collecting trajectories with varying iteration depths which incurs substantial upfront cost and typically yields limited diversity~\cite{deng2025supervised}.
(2) \textit{How to sample states}: standard sampling (e.g., uniform sampling) over the pool frequently selects states that are either {mastered} (deterministic high rewards) or beyond capability (deterministic failures) which leads to vanishing within-group reward dispersion, yielding ineffective updates and wastes computation~\cite{chen2026dragrpogrponeedsknow}.

To address these challenges, we propose \textbf{AceGRPO} (\textbf{A}daptive \textbf{C}urriculum \textbf{E}nhanced \textbf{GRPO}), an RL training framework that bootstraps an evolving pool of intermediate states to enable continuous, open-ended self-improvement. AceGRPO is built upon two tightly-coupled components: 
(1) To balance pool diversity with collection efficiency, we introduce an \textbf{Evolving Data Buffer} that continuously converts expensive executions into reusable step-wise training tasks. It treats every intermediate state (whether from failed debugging attempts or suboptimal solutions) as a valid starting point for a new single-step RL task.
(2) To spend execution budget on informative states, we design an \textbf{Adaptive Sampling} strategy that prioritizes states exhibiting both high reward variance (indicating uncertainty near the agent's capability frontier) and meaningful improvement headroom (avoiding mastered tasks with diminishing returns). By strategically sampling these high-value starting points, AceGRPO constructs an adaptive curriculum that concentrates computation on the agent’s evolving {Learning Zone}, preserving exploration diversity while improving training efficiency.

Empirically, AceGRPO improves both sample efficiency and overall performance. 
On OpenAI's MLE-Bench-Lite~\cite{chan2024mle}, Ace-30B (trained with AceGRPO) establishes a new state-of-the-art among open-source large models across all metrics, while also exceeding the performance of proprietary frontier models on specific indicators. Specifically, we observe a 12.13\% improvement in medal rate compared to Deepseek V3.2 and a 24.25\% gain relative to the untrained baseline. Extensive analysis and ablation experiments confirm the robustness of AceGRPO and underscore the critical role of the Evolving Data Buffer and Adaptive Sampling. And our code is available at \href{https://anonymous.4open.science/r/AceGRPO}{https://anonymous.4open.science/r/AceGRPO}.

Our key contributions are summarized as follows:
\begin{itemize}[leftmargin=1em,itemsep=0pt,parsep=0.2em,topsep=0.0em,partopsep=0.0em]
\item We propose \textbf{AceGRPO}, an adaptive RL framework that reformulates long-horizon MLE optimization as step-wise learning over an \textbf{Evolving Data Buffer}, enabling continuous self-evolution.

\item We introduce \textbf{Adaptive Sampling} guided by \textbf{Learnability Potential}. By serving as a proxy for gradient magnitude, this mechanism dynamically prioritizes tasks at the agent's learning frontier, ensuring exploration diversity while maximizing training efficiency.

\item On \textbf{MLE-Bench-Lite}, our Ace-30B achieves a \textbf{100\% valid submission rate} and shows stronger iterative optimization behavior than substantially larger frontier baselines.
\end{itemize}

\section{Related Work}

    \subsection{Autonomous Machine Learning Engineering}

    The evolution of LLM-driven agents facilitates autonomous Machine Learning Engineering (MLE), shifting focus from static QA to complex, iterative optimization. Foundational benchmarks like MLE-Bench~\cite{chan2024mle} and platforms like OpenHands~\cite{wang2024openhands} established the necessity for tight execution feedback loops. To navigate extended horizons, contemporary approaches predominantly rely on inference-time exploration strategies. Methods such as AIDE~\cite{jiang2025aide}, R\&D-Agent~\cite{yang2025r}, and ML-Master~\cite{mlmaster} utilize prompt-based planning, while AIRA~\cite{toledo2025aira}, AutoMLGen~\cite{du2025automlgen}, and FM Agent~\cite{li2025fm} incorporate evolutionary mechanisms for cross-branch knowledge sharing; others focus on tool optimization~\cite{sahney2025operand}, targeted initialization~\cite{nam2025mlestar} or context management~\cite{zhu2026toward}. However, these parameter-frozen approaches lack mechanisms to internalize learned strategies, leading to behavioral stagnation. While Reinforcement Learning (RL) offers a pathway to train capable solvers, applying it to long-horizon MLE is hindered by the prohibitive latency of full-trajectory feedback and the limited exploration offered by static datasets. Our work resolves this tension with AceGRPO, which bootstraps a dynamic task pool using a Learnability Potential to prioritize high-value intermediate states, ensuring sample-efficient self-evolution.

\subsection{Agentic Reinforcement Learning for LLMs}

The emergence of Agentic Reinforcement Learning (Agentic RL) marks a fundamental shift from treating LLMs as static text emitters to conceptualizing them as learnable policies embedded within sequential decision loops~\cite{cao2024survey}. This paradigm has catalyzed advances in "slow reasoning" capabilities, where models like DeepSeek-R1~\cite{guo2025deepseek} and o1-Coder~\cite{zhang2024o1} internalize deliberate thought processes and self-correction via reinforcement signals. In the domain of software engineering, AceCoder~\cite{zeng2025acecoder} and DeepCoder~\cite{deepcoder2025} progress from optimizing single-turn code generation to mastering complex, repository-scale workflows. Systems such as DeepSWE~\cite{deepswe2025}, SWE-RL~\cite{wei2025swe}, and Qwen3-Coder~\cite{yang2025qwen3} demonstrate that RL with execution-based feedback can effectively optimize agents for long-horizon debugging and feature implementation. Within MLE domains, ML-Agent~\cite{liu2025ml} extends this efficacy to automated machine learning pipelines. However, applying these methods to MLE remains challenged by the tension between the high latency of full-trajectory rollouts and the limited exploration of static datasets~\cite{zhao2025absolute, golubev2025training}. Our work addresses this bottleneck via AceGRPO, which employs an adaptive curriculum to dynamically prioritize high-potential intermediate states, ensuring efficient self-evolution without the computational prohibitive cost of redundant long-horizon exploration.

\section{Problem Formulation}
\label{sec:problem}

\begin{figure}
    \centering
    \vspace{-3em}
    \includegraphics[width=1.0\linewidth]{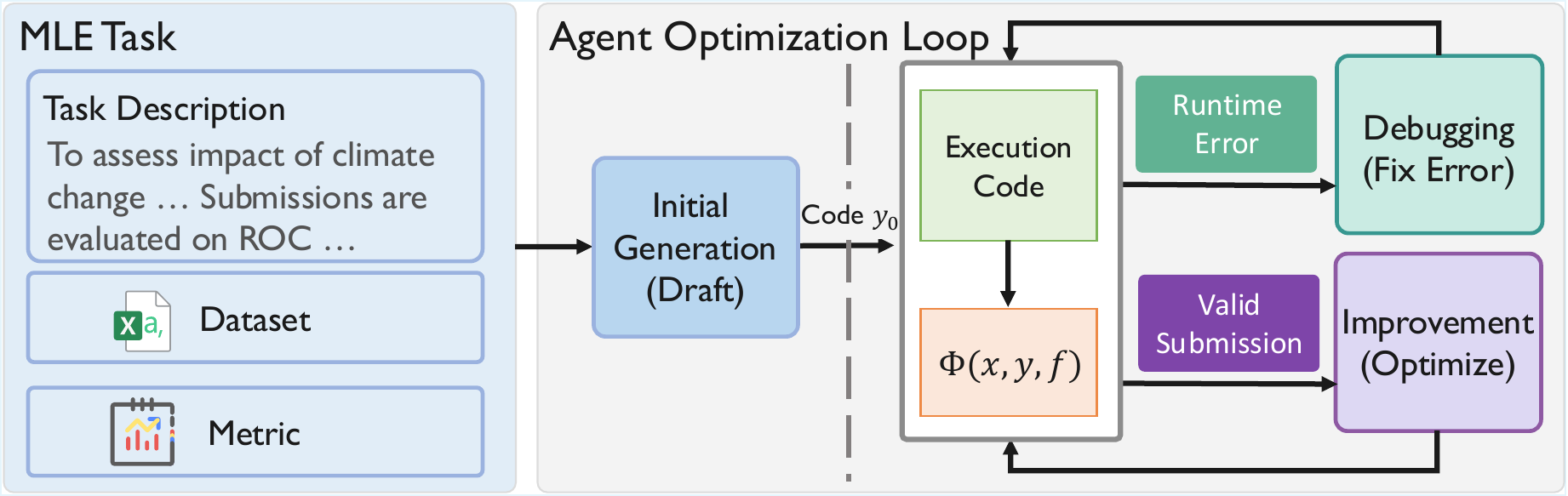}
    \caption{The Agent Optimization Loop for MLE. It is a continuous iteration over three distinct phases: Draft (initial generation), Debug (error correction), and Improve (metric optimization). The transitions are deterministically governed by the execution feedback, allowing the agent to refine its code solution and accumulate context through repeated interactions.}
    \label{fig:mle}
    \vspace{-2em}
\end{figure}

We formalize Machine Learning Engineering (MLE) as an iterative code optimization process. While the interaction naturally unfolds as a long-horizon agent-environment loop, we reformulate the learning objective as step-wise optimization over a dynamically evolving task distribution to address the fundamental efficiency constraints of the domain.

\subsection{MLE as Iterative Optimization with Task Expansion}


An MLE task instance is defined as a tuple $I=(\iota, \mathcal{D}, m)$, where $\iota$ is a natural-language task description, $\mathcal{D}$ represents the dataset, and $m: \mathcal{Y} \to \mathbb{R}$ is a metric function evaluating code $y \in \mathcal{Y}$.

The agent operates on a state (context) $x = (I, c, h) \in \mathcal{X}$, comprising the task instance $I$, the current code snapshot $c$, and an execution history $h = \{(y_k, f_k, s_k)\}_{k=0}^{|h|}$. 
Here, each entry records a past solution $y_k$, its execution feedback $f_k$, and the corresponding evaluation score $s_k$ derived from the metric $m$ (or a failure indicator if execution fails).
Given a context $x$, the policy $\pi_\theta(\cdot \mid x)$ generates a code update $y$. 
Depending on the state of $c$ and $h$, this action corresponds to three distinct operational phases, as illustrated in \cref{fig:mle}: \textit{Initial Generation} (when $c = \emptyset$), \textit{Debugging} (when the latest feedback in $h$ indicates failure), or \textit{Improvement} (when $c$ is valid but suboptimal).
Executing $y$ against the environment yields feedback $f = \mathrm{Exec}(x, y)$ and a scalar reward $r = \mathcal{R}(x, y)$.

Crucially, rather than treating the interaction as a static dataset, we model it as a dynamic expansion process. Each execution deterministically spawns a new derivative state via a transition operator $\Phi$:
\begin{equation}
    x' = \Phi(x, y, f) = (I, y, h \cup \{(y, f, r)\}).
\end{equation}
Let $\mathcal{B}_t \subset \mathcal{X}$ denote the evolving context buffer at iteration $t$. The buffer expands recursively as $\mathcal{B}_{t+1} \leftarrow \mathcal{B}_t \cup \{ \Phi(x, y, f) \}$. 
Since optimizing the full long-horizon trajectory return is intractable due to sparse feedback and high latency, we actuate a step-wise optimization as a substitute. 
We train the policy to maximize the expected immediate reward over an adaptive sampling distribution $\mathcal{Q}_t$ defined on the current buffer $\mathcal{B}_t$:
\begin{equation}
    J_t(\theta) = \mathbb{E}_{x \sim \mathcal{Q}_t} \left[ \mathbb{E}_{y \sim \pi_\theta(\cdot \mid x)} \big[ \mathcal{R}(x, y) \big] \right].
    \label{eq:objective}
\end{equation}
This formulation reframes MLE from standard trajectory optimization into \textit{active context selection}: the core challenge shifts to designing $\mathcal{Q}_t$ to strictly prioritize states where the gradient signal maximizes the agent's self-evolution per unit of execution cost.

\subsection{Reinforcement Learning for LLMs}

For a sampled context $x\sim\mathcal{Q}_t$, GRPO draws a group $\{y_i\}_{i=1}^G$ from $\pi_{\theta_{\text{old}}}(\cdot\mid x)$ and evaluates $r_i=\mathcal{R}(x,y_i)$. Define

\vspace{-1em}
\begin{equation}
\begingroup
\small
\begin{gathered}
\mu(x) = \frac{1}{G}\sum_{i=1}^G r_i,
\sigma(x) = \sqrt{\frac{1}{G}\sum_{i=1}^G \big(r_i-\mu(x)\big)^2}, 
\hat{A}_i(x) = \frac{r_i-\mu(x)}{\sigma(x)+\varepsilon}.
\end{gathered}
\endgroup
\end{equation}

Let $\rho_i(\theta)=\pi_\theta(y_i\mid x)/\pi_{\theta_{\text{old}}}(y_i\mid x)$. With clipping $\delta$ and KL regularization to a reference policy $\pi_{\mathrm{ref}}$ weighted by $\beta$, GRPO maximizes
\begin{equation}
\begingroup
\tiny
\begin{aligned}
\mathcal{L}_{\mathrm{GRPO}}(\theta)
=&\ \mathbb{E}_{x\sim \mathcal{Q}_t}\Bigg[
\frac{1}{G}\sum_{i=1}^G
\min\!\Big(
\rho_i(\theta)\hat{A}_i(x),\,
\mathrm{clip}(\rho_i(\theta),1-\delta,1+\delta)\hat{A}_i(x)
\Big) 
-\beta\,\mathrm{KL}\!\left(\pi_\theta(\cdot\mid x)\,\|\,\pi_{\mathrm{ref}}(\cdot\mid x)\right)
\Bigg].
\end{aligned}
\endgroup
\end{equation}

\subsection{Challenges in RL for Dynamic MLE Tasks}
\paragraph{Prohibitive feedback latency.}
Unlike fast-verification domains, evaluating $\mathcal{R}(x,y)$ often requires running full training MLE pipelines, making multi-step online trajectory collection impractical at scale. This motivates the step-wise objective in Eq.~\eqref{eq:objective} and shifts the bottleneck to selecting informative contexts $x$ under a fixed execution budget.

\paragraph{Non-stationary expanding pools and variance collapse.}
Because $\mathcal{T}_t$ grows via $\Phi$, naive sampling (e.g., uniform over $\mathcal{T}_t$) is dominated by contexts that quickly become either (i) \emph{mastered}, where $r_i\approx r_{\max}$ for all $i$, or (ii) \emph{beyond-frontier}, where $r_i\approx r_{\min}$ for all $i$. In both regimes, $r_i-\mu(x)\to 0$, hence $\hat{A}_i(x)\to 0$, and the GRPO update loses signal:
\begin{equation}
\nabla_\theta \mathcal{L}_{\mathrm{GRPO}}(\theta)
\;\propto\;
\mathbb{E}\big[\nabla_\theta \log \pi_\theta(y_i\mid x)\ \hat{A}_i(x)\big]
\;\approx\;0.
\end{equation}
Therefore, uniform sampling wastes compute on ``empty'' steps. This motivates \textbf{AceGRPO}, which replaces naive $\mathcal{Q}_t$ with an adaptive curriculum that targets contexts in the agent's learning zone.

\section{AceGRPO}
\label{sec:mathod}

\begin{figure}
    \vspace{-3em}
    \centering
    \includegraphics[width=1.0\linewidth]{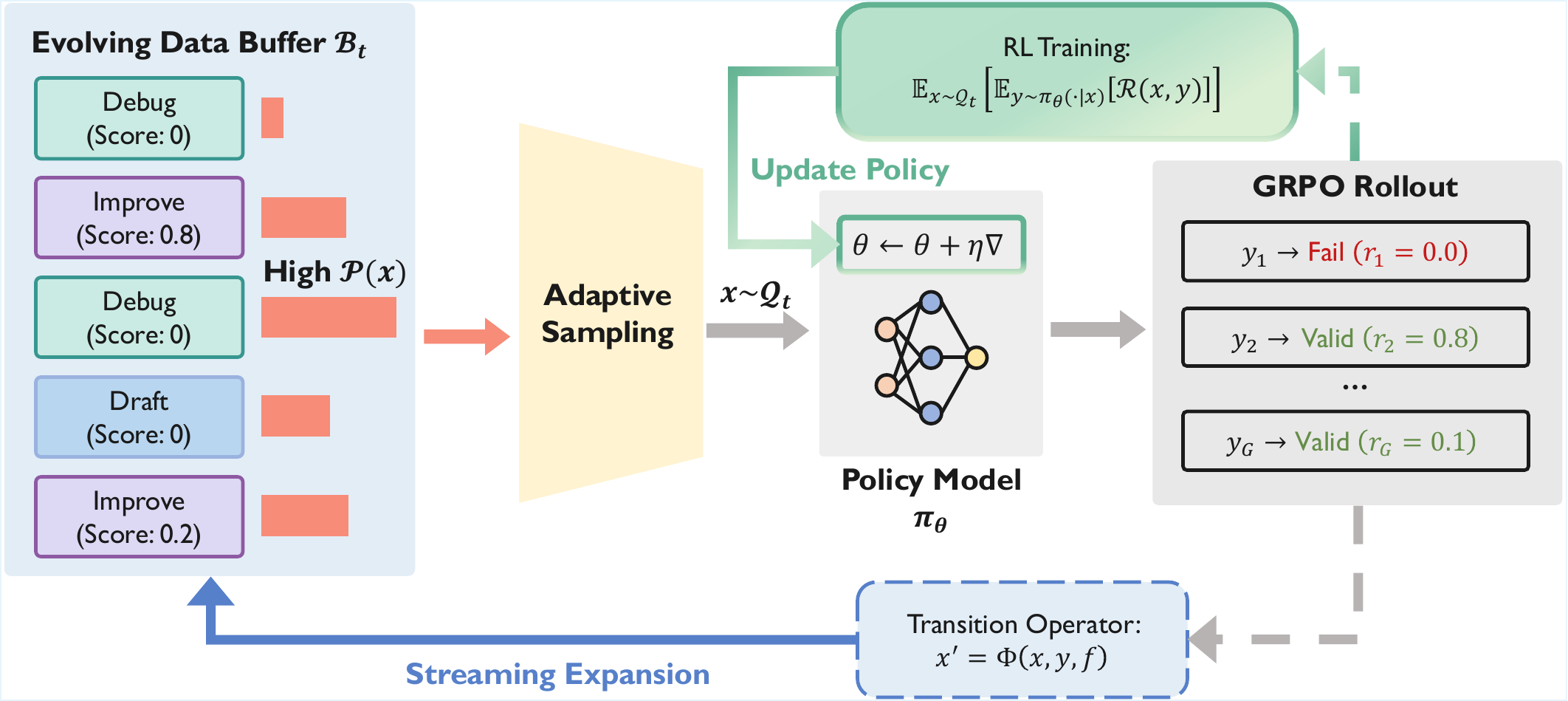}
    \caption{The AceGRPO framework. MLE is step-wise optimization over a dynamic task distribution. The method features an Evolving Data Buffer that accumulates intermediate states (Draft, Debug, Improve) via Streaming Expansion. An Adaptive Sampler selects high-potential tasks to maximize the gradient signal for GRPO training, enabling efficient self-improvement on long-horizon challenges.}
    \label{fig:overview}
    \vspace{-1.5em}
\end{figure}

To address the prohibitive execution latency and the stage-varying ability requirements in long-horizon Machine Learning Engineering (MLE), we propose \textbf{AceGRPO}, an RL framework built upon two tightly-coupled components:
(1) an \textbf{Evolving Data Buffer} that continuously converts expensive executions into reusable step-wise training tasks, and
(2) an \textbf{Adaptive Sampling} strategy that adaptively allocates the limited execution budget to the agent's current learning frontier and the capability emphasized at the current optimization stage.

\subsection{Evolving Data Buffer: Streaming Construction of Step-wise Training Tasks}
\label{sec:buffer}

While MLE interactions naturally unfold as long trajectories, optimizing them via full-trajectory RL is intractable due to high feedback latency and prohibitive execution costs.
AceGRPO operationalizes the step-wise objective in \Cref{eq:objective} by maintaining an \textbf{evolving data buffer} $\mathcal{B}_t$. 
This buffer serves as a dynamic context pool, allowing the agent to continuously repurpose expensive execution traces into distinct, replayable optimization steps. 
In practice, the outer-loop sampling distribution $\mathcal{Q}_t$ is defined directly over this buffer, i.e., $\mathrm{supp}(\mathcal{Q}_t) \subseteq \mathcal{B}_t$.

\textbf{Buffer content.}
Each item in $\mathcal{B}_t$ represents an intermediate state $x = (I, c, h)$, fully encapsulating the problem description, current code snapshot, and execution history. 
To facilitate adaptive sampling, we augment each state with lightweight metadata:
(i) a \emph{task type} $\kappa(x) \in \{\textsc{Draft}, \textsc{Debug}, \textsc{Improve}\}$, and
(ii) a \emph{learnability potential} $\mathcal{P}(x)$ (detailed in \Cref{sec:sampling}). 
Here, \textsc{Draft} denotes solving from the task description, \textsc{Debug} denotes repairing an invalid execution using error feedback, and \textsc{Improve} denotes refining a valid solution using its previous score as a baseline. 
The $\mathcal{P}(x)$ is initialized to a default value $\mathcal{P}_{\text{init}}$ upon insertion 
and it is updated only when $x$ is sampled from the buffer and used as the initial state of a new rollout. 
In this case, $\mathcal{P}(x)$ is updated using the reward produced by the most recent rollout.

\textbf{Streaming expansion.} 
At training step $t$, we sample a batch of starting states $x \sim \mathcal{Q}_t(\mathcal{B}_t)$ and generate a group of $G$ candidate actions $\{y_i\}_{i=1}^G \sim \pi_\theta(\cdot \mid x)$.
Executing each $y_i$ yields feedback $f_i$ (e.g., error logs or leaderboards) and a raw execution score $s_i$.
Crucially, strictly following the transition operator defined in \Cref{sec:problem}, each execution spawns a new derivative state $x'_i = \Phi(x, y_i, f_i)$, which is immediately appended to $\mathcal{B}_t$.
The task type $\kappa(x'_i)$ is assigned logically based on the outcome: failed runs generate \textsc{Debug} states enriched with tracebacks, while successful runs produce \textsc{Improve} states containing the new score as a baseline.

This mechanism transforms every execution into two distinct assets: a gradient signal for the current policy update and a novel starting point for future curriculum learning.
By expanding $\mathcal{B}_t$ on-the-fly with the agent's own outputs, AceGRPO ensures that the training distribution continuously shifts towards the agent's evolving frontier, preserving fresh policy-generated diversity without the cost of redundant full-trajectory rollouts.

\textbf{Reward shaping.} 
We adopt the \textit{HumanRank score}~\cite{qiang2025mle} as the unified metric.
Let $p$ be the rank of a submission among $N$ human participants. 
The normalized score is $s = 1 - \frac{p}{N}$. 
Invalid submissions are assigned $s = -1$.
We define the immediate reward $R(x, y)$ as a mixture of absolute performance and relative improvement over the state's previous baseline $s_p(x)$.
For \textsc{Draft} and \textsc{Debug} tasks, $s_p(x) = 0$; for \textsc{Improve} tasks, $s_p(x)$ is the previous valid score inherited from the parent state.
\begin{equation}
\begingroup
R(x,y)=
\begin{cases}
0, & s=-1,\\
(1-\alpha)s+\alpha\max\!\left(0,\dfrac{s-s_{\mathrm p}(x)}{1-s_{\mathrm p}(x)+\epsilon}\right), & \text{otherwise},
\end{cases}
\label{eq:shaped_reward}
\endgroup
\end{equation}
where $\alpha \in [0, 1]$ balances the terms and $\epsilon$ ensures numerical stability.
This shaping strategy creates a stage-consistent incentive: early training focuses on achieving valid solutions (via the absolute term $s$), while later stages are explicitly driven to exploit the diminishing improvement headroom (via the relative term).

\textbf{Updating potential from GRPO group outcomes.} 
When a state $x$ is sampled, the GRPO rollout produces a reward set $\mathbf{r}_x^{(t)} = \{R(x, y_1), \ldots, R(x, y_G)\}$.
We utilize the within-group statistics of $\mathbf{r}_x^{(t)}$ to update $\mathcal{P}(x)$, as this variance serves as a direct proxy for the learning signal's magnitude (see \Cref{sec:sampling}).
For any newly spawned state $x'$, we set $\mathcal{P}(x') = \mathcal{P}_{\text{init}}$ pending its first execution.

\subsection{Adaptive Sampling: Dynamic Prioritization via Learnability Potential}
\label{sec:sampling}

The buffer $\mathcal{B}_t$ provides a rich and expanding substrate of training states. 
However, each sampled starting state triggers expensive executions, 
and the effectiveness of a GRPO update depends critically on whether the sampled state induces non-trivial within-group reward variation. 
Standard uniform sampling is inefficient because the buffer inevitably accumulates states that are either \textit{mastered} (yielding deterministic high rewards) or \textit{beyond capability} (yielding deterministic failures). Both regimes produce vanishing within-group advantage variance ($\hat{A}_i \approx 0$), resulting in wasted computation.
Consequently, we explicitly design the outer-loop sampling distribution $\mathcal{Q}_t$ (\Cref{eq:objective}) as a strategic resource allocation mechanism for allocating execution budget.

We construct $\mathcal{Q}_t$ as a potential-weighted distribution over task types, combined with lightweight diversity terms that preserve long-term coverage.

\textbf{Learnability Potential as a Gradient Proxy.} 
We introduce \textit{Learnability Potential}, $\mathcal{P}(x)$, to quantify the expected informativeness of a state based on its most recent execution outcomes.
Since the magnitude of the GRPO update is proportional to the advantage variance, the dispersion of $\mathbf{r}_x^{(t)}$ serves as a direct proxy for the learning signal. We define $\mathcal{P}(x)$ as:
\begin{equation}
\mathcal{P}(x) =
\underbrace{\mathrm{clip}\!\bigl(\sigma(\mathbf{r}_x^{(t)}), 0, \delta_{\max}\bigr)}_{\text{Uncertainty}}
\cdot
\underbrace{\mathrm{clip}\!\bigl(1-\mu(\mathbf{r}_x^{(t)}), 0, 1\bigr)}_{\text{Headroom}},
\label{eq:potential}
\end{equation}
where $\mu(\cdot)$ and $\sigma(\cdot)$ denote the mean and standard deviation, respectively. 
Here, $\delta_{\max}$ caps extreme variance to prevent instability. 
The \textit{Uncertainty} term targets states in the ``learning zone'' where outcomes are non-deterministic and within-group variance is non-zero,
mitigating GRPO gradient collapse. 
The \textit{Headroom} term only suppresses saturated high-mean states; it is not intended to penalize low-mean states by itself. Instead, the multiplicative form makes uncertainty a gate on headroom: deterministic failures receive low potential because $\sigma(\mathbf{r}_x^{(t)})\approx 0$, while states receive high potential only when they exhibit both non-trivial reward dispersion and remaining improvement space.
For any newly inserted state $x'$ not yet executed, we initialize $\mathcal{P}(x') = \mathcal{P}_{\text{init}}$.

\textbf{Smooth curriculum with recency.} 
Directly sampling proportional to $\mathcal{P}(x)$ can lead to distribution collapse, where the agent overfits to a narrow set of high-variance tasks. 
To ensure robustness, AceGRPO uses a temperature-controlled sampler that is exploratory when the buffer is small and becomes more selective as the buffer grows:
\begin{equation}
\tau_t=\tau_0\max\!\left(\epsilon_\tau,1-\frac{|\mathcal{B}_t|}{N_{\mathrm{sat}}}\right)
\label{eq:temperature}
\end{equation}
We also down-weight states that were sampled very recently:
\begin{equation}
D_t(x)=\min\!\left(1,\frac{t-t_{\mathrm{last}}(x)}{\Delta}\right),
\label{eq:recency}
\end{equation}
where $t_{\mathrm{last}}(x)$ is the last step when $x$ was selected.
This simple recency factor plays the same role as a refractory period, preventing a few states from monopolizing the rollout budget.

\begin{table}[!t]
\vspace{-3em}
\caption{Main evaluation results on MLE-Bench-Lite. All results are reported as mean $\pm$ standard deviation. The best open-source model performance is marked \textbf{bold}. Qwen3-235B-2507 refers to Qwen3-235B-A22B-Thinking-2507, and Qwen3-30B-2507 refers to Qwen3-30B-A3B-Thinking-2507. Except for the HumanRank Score multiplied by 100 for readability (defined in $[0, 1]$), all other metrics are reported as percentages.}
\label{tab:main_results}
\centering
\begin{small}
\setlength{\tabcolsep}{4.2pt}
\renewcommand{\arraystretch}{1.25}
\resizebox{\linewidth}{!}{
\begin{tabular}{lccccc|cc}
\toprule
Model 
& \makecell[c]{Valid\\Submission}
& \makecell[c]{Above\\Median}
& Bronze
& Silver
& Gold
& \makecell[c]{Any\\Medal}
& \makecell[c]{HumanRank \\ Score} \\
\midrule
Claude-4.5-Sonnet
& 100.00 {\scriptsize $\pm$ 0.00}
& 80.30  {\scriptsize $\pm$ 2.14}
& 7.58   {\scriptsize $\pm$ 5.67}
& 16.67  {\scriptsize $\pm$ 5.67}
& 36.36  {\scriptsize $\pm$ 3.71}
& 60.61  {\scriptsize $\pm$ 5.67}
& 75.86  {\scriptsize $\pm$ 1.88} \\

Gemini-3-Pro 
& 100.00 {\scriptsize $\pm$ 0.00}
& 84.85  {\scriptsize $\pm$ 4.29}
& 4.55   {\scriptsize $\pm$ 3.71}
& 15.15  {\scriptsize $\pm$ 4.29}
& 39.39  {\scriptsize $\pm$ 2.14}
& 59.09  {\scriptsize $\pm$ 3.71}
& 79.18  {\scriptsize $\pm$ 2.40} \\

GPT-5.2 
& 95.45  {\scriptsize $\pm$ 3.71}
& 71.21  {\scriptsize $\pm$ 4.29}
& 10.61  {\scriptsize $\pm$ 4.29}
& 16.67  {\scriptsize $\pm$ 2.14}
& 28.79  {\scriptsize $\pm$ 2.14}
& 56.06  {\scriptsize $\pm$ 4.29}
& 71.05  {\scriptsize $\pm$ 0.91} \\

\midrule
GLM-4.6
& 90.91  {\scriptsize $\pm$ 3.71}
& 45.45  {\scriptsize $\pm$ 3.71}
& 6.06   {\scriptsize $\pm$ 4.29}
& 3.03   {\scriptsize $\pm$ 2.14}
& 21.21  {\scriptsize $\pm$ 4.29}
& 30.30  {\scriptsize $\pm$ 3.71}
& 45.45  {\scriptsize $\pm$ 7.73} \\


DeepSeek-V3.2 
& 98.48  {\scriptsize $\pm$ 2.14}
& 71.21  {\scriptsize $\pm$ 2.14}
& 7.58   {\scriptsize $\pm$ 2.14}
& 4.55   {\scriptsize $\pm$ 3.71}
& 27.27  {\scriptsize $\pm$ 3.71}
& 39.39  {\scriptsize $\pm$ 2.14}
& 65.92  {\scriptsize $\pm$ 1.62} \\

Qwen3-235B-2507
& 92.42  {\scriptsize $\pm$ 2.14}
& 69.70  {\scriptsize $\pm$ 7.73}
& 6.06   {\scriptsize $\pm$ 5.67}
& 7.58   {\scriptsize $\pm$ 4.29}
& 24.24  {\scriptsize $\pm$ 4.29}
& 37.88  {\scriptsize $\pm$ 10.71}
& 69.70  {\scriptsize $\pm$ 7.73} \\

\midrule
Qwen3-30B-2507
& 84.85  {\scriptsize $\pm$ 2.14}
& 50.00  {\scriptsize $\pm$ 3.71}
& 1.52   {\scriptsize $\pm$ 2.14}
& 7.58   {\scriptsize $\pm$ 2.14}
& 18.18  {\scriptsize $\pm$ 3.71}
& 27.27  {\scriptsize $\pm$ 6.43}
& 48.75  {\scriptsize $\pm$ 3.45} \\

\textbf{Ace-30B} 
& \textbf{100.00} {\scriptsize $\pm$ \textbf{0.00}}
& \textbf{78.79}  {\scriptsize $\pm$ \textbf{5.67}}
& \textbf{12.12}  {\scriptsize $\pm$ \textbf{4.29}}
& \textbf{7.58}   {\scriptsize $\pm$ \textbf{2.14}}
& \textbf{31.82}  {\scriptsize $\pm$ \textbf{3.71}}
& \textbf{51.52}  {\scriptsize $\pm$ \textbf{8.57}}
& \textbf{71.14}  {\scriptsize $\pm$ \textbf{4.08}} \\

\bottomrule
\end{tabular}
}
\end{small}
\vspace{-1em}
\end{table}

\textbf{Final sampling distribution.} 
Combining the potential-based priority, temperature, and recency factor, the final probability of sampling state $x$ at iteration $t$ is given by:
\begin{equation} 
\mathcal{Q}_t(x) = \frac{1}{Z_t}
\exp\!\left(\frac{\mathcal{P}(x)}{\tau_t}\right)
D_t(x),
\label{eq:final_qt}
\end{equation}
where $Z_t$ is the partition function normalizing the distribution over $\mathcal{B}_t$. 
This construction dynamically aligns the execution budget with the agent's evolving competency while preserving necessary exploration diversity.

\subsection{Asynchronous Training Architecture}

To handle the latency of MLE tasks, we implement AceGRPO using a decoupled, asynchronous architecture. 
The system separates \textbf{Rollout Workers} from \textbf{Learner Actors}. 
Workers continuously sample starting states $x \sim \mathcal{Q}_t$ from the evolving buffer $\mathcal{B}_t$, execute actions, and asynchronously perform the streaming expansion of $\mathcal{B}_t$ with new states. 
Simultaneously, Learner Actors consume the collected interaction data to perform GRPO updates. 
This parallelization ensures that the curriculum statistics and buffer content are refreshed in real-time without blocking the optimization loop, effectively maximizing the throughput of valid learning samples.

\section{Experiments}
\label{sec:exp}

\subsection{Experimental Setup}

\textbf{Training settings.}
In our experiments, we fine-tuned Qwen3-30B-A3B-Thinking-2507~\cite{yang2025qwen3} using our AceGRPO to obtain Ace-30B.
We use MLE-Dojo~\cite{qiang2025mle} to construct our training dataset, removing the 68 tasks that overlap with MLE-Bench~\cite{chan2024mle}. The final training dataset consists of 134 machine learning competition tasks.
The reinforcement learning training runs for 400 steps with a rollout batch size of 8 tasks per iteration, 8 rollouts per task, and a global training batch size of 64. 
Additional training details are provided in Appendix~\ref{app:training_settings}. And prompts are in Appendix~\ref{app:prompts}

\textbf{Evaluation settings.}
We evaluate our trained models on MLE-Bench-Lite, a 22-task Kaggle-style subset of MLE-Bench~\cite{chan2024mle}.
Each competition presents a realistic Kaggle-style challenge where the agent must analyze the problem, train models, and submit predictions—all within a 12-hour time budget per task. 
Following MLE-Bench, we measure performance using the following metrics: \textit{valid submission}, \textit{above-median}, \textit{bronze}, \textit{silver}, \textit{gold}, and \textit{any medal}. Additionally, we report the \textit{HumanRank score} from MLE-Dojo~\cite{qiang2025mle}. Details about metrics and MLE-Bench are provided in \Cref{app:mle_bench}. 
We compare our approach against several strong baselines: (1) proprietary frontier models including Claude-4.5-Sonnet~\cite{Anthropic2025Claude4}, Gemini-3-Pro~\cite{Google2025Gemini3}, and GPT-5.2~\cite{OpenAI2025GPT5}; (2) open-source models including DeepSeek-V3.2~\cite{liu2025deepseek}, GLM 4.6~\cite{zeng2025glm}, and Qwen3-235B-A22B-Thinking-2507~\cite{yang2025qwen3}; and (3) the base Qwen3-30B-A3B-Thinking-2507 without RL training. We enable the thinking mode of all the baseline models and our Ace-30B in the evaluation. 
To ensure statistical reliability, we repeat each evaluation run three times and report mean performance with standard deviation.

\subsection{Main Results}


\textbf{AceGRPO is an effective reinforcement learning method that substantially improves end-to-end MLE performance.} 
Table~\ref{tab:main_results} presents a comprehensive comparison with proprietary frontier and large open-source LLMs on the MLE-Bench-Lite. 
As it shows, Ace-30B consistently outperforms the baseline Qwen3-30B-A3B-Thinking model in all major evaluation metrics. 
In particular, the Any Medal rate increases by 24.25\% (from 27.27\% to 51.52\%), and the HumanRank score improves by 22.39\% (from 48.75 to 71.14). 
Notably, Ace-30B also achieves a \textbf{100\% Valid Submission rate}, matching the strongest proprietary LLM (Claude-4.5-Sonnet) and indicating substantially improved robustness in long-horizon MLE tasks. 
These consistent gains across multiple metrics demonstrate that AceGRPO yields effective reinforcement learning updates, enabling LLMs to internalize machine learning engineering strategies beyond inference-time prompting.

\begin{figure}[!t]
    \centering
    \vspace{-3em}
    \includegraphics[width=1.0\linewidth]{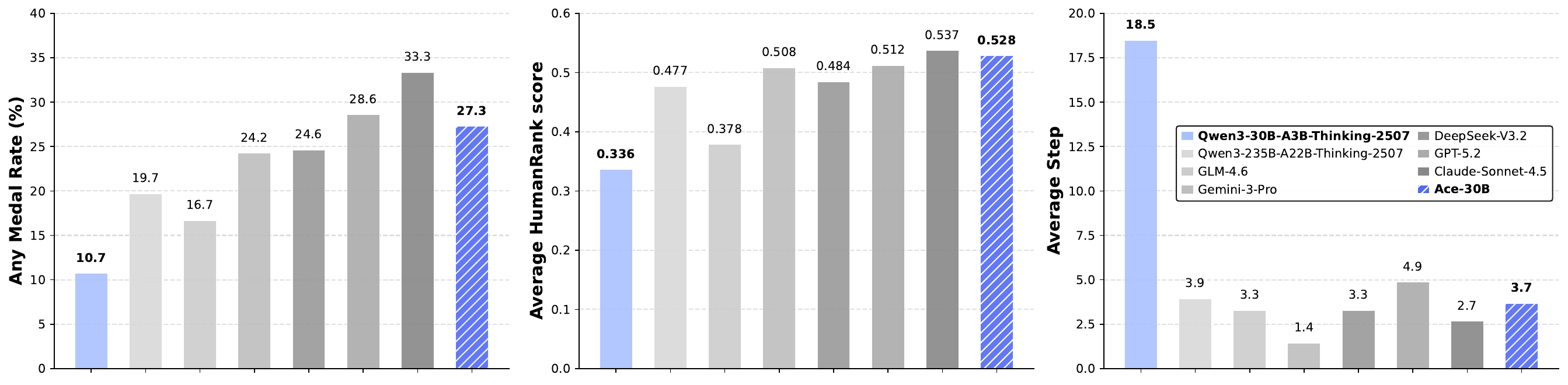}
    \caption{Performance comparison on first valid submissions. We report: (left) medal rate on the first valid submission, (middle) average HumanRank score of the first valid submission, and (right) average number of agent steps needed to produce the first valid submission. Lower steps indicate faster generation of valid solutions.}
    \label{fig:first_submission}
     \vspace{-1.5em}
\end{figure}

\textbf{AceGRPO approaches the performance of proprietary frontier LLMs while outperforming larger open-source baselines.}
As shown in Table~\ref{tab:main_results}, Ace-30B attains aggregated performance that is competitive with proprietary frontier models on both Any Medal rate and HumanRank score, despite being trained from a substantially smaller open-source LLM. In particular, Ace-30B reaches a 51.52\% Any Medal rate and a 71.14\% HumanRank score, matching the level of GPT-5.2 (56.06\% Any Medal; 71.05 HumanRank score) and even slightly surpassing it on Human Rank. Meanwhile, Ace-30B clearly outperforms larger open-source baselines such as DeepSeek-V3.2 (39.39\% Any Medal; 65.92 HumanRank score) and Qwen3-235B-2507 (37.88\% Any Medal; 69.70 HumanRank score). These results suggest that AceGRPO closes a substantial portion of the gap to closed-source frontier LLMs and delivers stronger performance than much larger open-source models under realistic MLE evaluation settings.

\begin{figure}[t]
\begin{minipage}{0.4\linewidth}
    \centering
    \includegraphics[width=\linewidth]{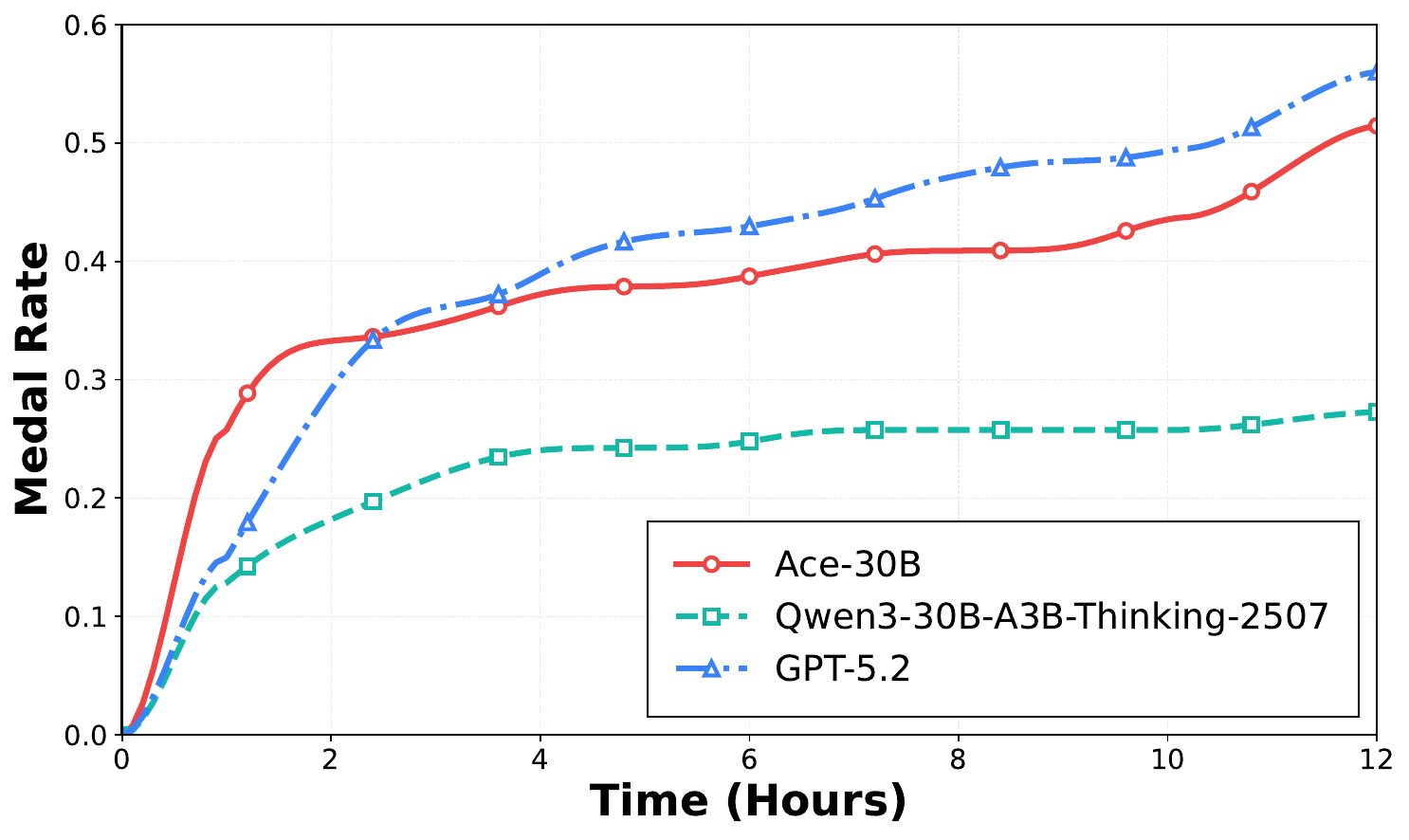}
    \caption{Medal rate evolution over time.}
    \label{fig:medal_rate_time}
\end{minipage}
\hspace{1em}
\begin{minipage}{0.55\linewidth}
    \centering
    \captionof{table}{Ablation and training-baseline study on MLE-Bench-Lite. Component ablations use the same 400-step RL budget. $\Delta$ denotes the improvement in Any Medal rate over the base Qwen3-30B model.}
    \resizebox{\linewidth}{!}{
    \begin{tabular}{lccc}
    \toprule
    Method & Any Medal & HumanRank & $\Delta$ \\
    \midrule
     Qwen3-30B-A3B-2507 & 27.27\% & 48.75 & - \\
     SFT & 36.36\% & 57.80 & +9.09\% \\
     Vanilla GRPO & 34.85\% & 55.90 & +7.58\% \\
    \midrule
    \textbf{AceGRPO} & \textbf{51.52\%} & \textbf{71.14} & \textbf{+24.25\%} \\
    \quad w/o Evolving Data Buffer & 45.45\% & 65.80 & +18.18\% \\
    \quad w/o Adaptive Sampling & 42.73\% & 62.70 & +15.46\% \\
    \bottomrule
    \end{tabular}}
    \label{tab:ablation}
\end{minipage}
\vspace{-2em}
\end{figure}

\textbf{AceGRPO enables sustained self-evolution rather than one-shot performance gains.}
Figure~\ref{fig:medal_rate_time} plots the Any Medal rate over time on MLE-Bench-Lite for different LLMs.
Whereas Qwen3-30B-A3B-Thinking largely plateaus within the first six hours, Ace-30B continues to improve steadily throughout the entire run, consistently surpassing the base model and progressively narrowing the gap to GPT-5.2.
This persistent upward trajectory suggests that AceGRPO facilitates sustained self-improvement over extended horizons, rather than yielding isolated or brittle gains, supporting its effectiveness for MLE-style tasks.

\subsection{Ablation Study}
We compare AceGRPO with an SFT baseline, Vanilla GRPO, and two component ablations that remove the Evolving Data Buffer or Adaptive Sampling. 
Additional hyperparameter sensitivity analysis is reported in \Cref{app:hyperparameter_ablation}. 
We highlight two main conclusions below.

\textbf{AceGRPO is more effective than SFT from successful trajectories.} 
We train an SFT baseline. The SFT data comes from the AceGRPO training rollout trajectories, and contains 14K samples distilled from \textsc{Draft}, \textsc{Debug}, and \textsc{Improve} states. 
\Cref{tab:ablation} shows  SFT improves Any Medal from 27.27\% to 36.36\%, showing that successful trajectories provide useful supervision. 
However, AceGRPO reaches 51.52\% Any Medal and 71.14 HumanRank, substantially outperforming SFT by 15.16\% in Any Medal and 13.34\% in HumanRank.
This demonstrates that AceGRPO's online execution-grounded training provides a stronger signal than static imitation, especially because the policy learns from its own intermediate states, failures, and recovery attempts.

\textbf{The evolving data buffer and adaptive sampling are both essential to AceGRPO.}
All RL variants are trained with a matched 400-step budget.
As shown in \Cref{tab:ablation}, both components are critical to AceGRPO's performance. 
Removing the \textbf{Evolving Data Buffer} reduces the Any Medal rate by 6.07\%, highlighting the importance of streaming task expansion. Without dynamically repurposing intermediate execution traces into new training tasks, the agent is confined to repetitive trajectories and cannot self-evolve beyond the static dataset distribution in the absence of on-policy interaction.
Removing \textbf{Adaptive Sampling} causes an even larger 8.79\% drop in Any Medal rate. This underscores the value of Adaptive Sampling and our proposed Learnability Potential Function: as the task pool grows, naive sampling becomes increasingly inefficient because the learning signal is diluted by already-mastered or effectively intractable tasks. 
\textbf{Vanilla GRPO}, which lacks both mechanisms, obtains only a 7.58\% improvement over the base LLM.
These results show that AceGRPO's improvement is not simply due to applying GRPO, but comes from coupling online state evolution with adaptive sampling of informative training states. 

\vspace{-1em}
\subsection{Analysis}

\textbf{AceGRPO significantly improves both the quality and efficiency of early-stage solutions.} Figure~\ref{fig:first_submission} compares different LLMs based on their first valid submission, and reports three complementary metrics: Any Medal rate, average HumanRank score, and the number of steps required to reach the first valid solution. 
These metrics jointly characterize not only the quality of the initial feasible solutions but also the efficiency with which an LLM escapes invalid states. 
Ace-30B achieves a substantially higher medal rate on the first valid submission, improving from 10.71\% for the base Qwen3-30B-A3B-Thinking model to 27.27\%. 
At the same time, it significantly reduces the average number of steps required to reach a valid submission from 18.48 down to 3.67. 
This indicates that AceGRPO enables LLMs to generate valid and competitive solutions much earlier in the optimization process, rather than relying on prolonged trial-and-error to recover from failures. 
Notably, Ace-30B’s first-submission performance is competitive with larger proprietary frontier LLMs such as Claude-4.5-Sonnet and GPT-5.2, both in terms of solution quality and solution emergence speed. 
This suggests that the improvements introduced by AceGRPO are not limited to final performance but also manifest strongly at early stages of interaction, where efficient exploration and rapid stabilization are critical~\cite{zhu2026toward}. 
Overall, Figure~\ref{fig:first_submission} demonstrates that AceGRPO enhances early-stage robustness and efficiency, enabling LLM-based MLE agents to transition more reliably and quickly from invalid states to meaningful optimization trajectories.

\begin{figure}[!t]
    \centering
    \vspace{-3em}
    \includegraphics[width=\linewidth]{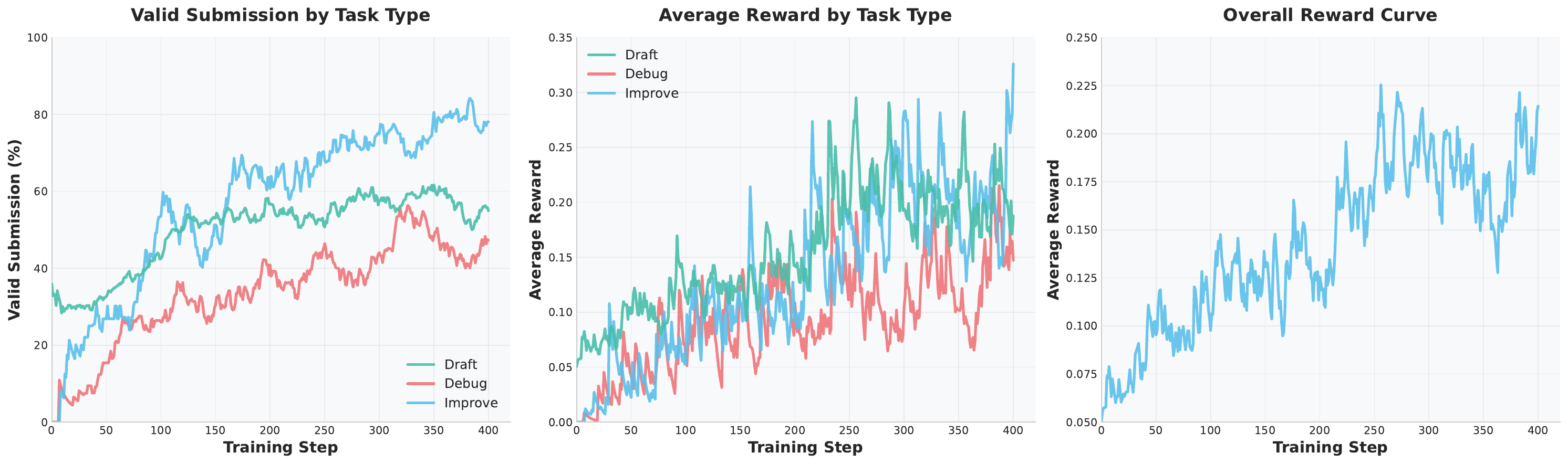}
    \caption{Training dynamics of rewards curve and performance of valid submission}
    \label{fig:reward}
     \vspace{-1.5em}
\end{figure}

\textbf{AceGRPO exhibits consistent and effective training dynamics.} \Cref{fig:reward} illustrates the training dynamics of AceGRPO. \Cref{fig:reward} (Left) depicts the evolution of the valid submission rate for the Draft, Debug, and Improve tasks. The consistent upward trend across all three tasks suggests steady gains in the LLM’s baseline capabilities.  \Cref{fig:reward} (Middle) presents the reward trajectories for these tasks throughout the training process. The concurrent rise in rewards demonstrates the model's enhanced proficiency in generating high-quality code and solving machine learning problems.
\Cref{fig:reward} (Right) illustrates the overall reward curve aggregated across all tasks, which exhibits a increase.
Taken together, these results indicate that the LLM improves continuously during training, as evidenced by rising valid submission rates and rewards across all task types. 



\section{Conclusion}

In this work, we presented \textbf{AceGRPO}, a reinforcement learning framework designed to bridge the gap between transient inference-time search and persistent policy internalization for long-horizon autonomous Machine Learning Engineering. To address the challenges of prohibitive feedback latency, AceGRPO reformulates the learning process as step-wise optimization over a dynamically evolving task distribution. By treating every intermediate execution as a potential curriculum entry and strictly prioritizing tasks within the agent's "Learning Zone" via the \textit{Learnability Potential} function, AceGRPO ensures that limited computational budgets are invested where gradient signals are most informative and prevents the dilution of the learning signal.

Empirical results on MLE-Bench-Lite validate the efficacy of AceGRPO. Our Ace-30B trained with AceGRPO achieves 100\% Valid Submission, 51.52\% medal rate and 0.7114 HumanRank score, outperforming all the large open-source LLMs. Crucially, AceGRPO enables a 30B model to exhibit optimization behaviors and final performance comparable to proprietary frontier LLMs like GPT-5.2. Overall, AceGRPO paves the way for future research into efficiently training self-evolving agents capable of autonomous handling long-horizon task like Machine Learning Engineering.

\bibliographystyle{unsrtnat}
\bibliography{ref}


\newpage
\appendix
\section{Implementation Details}
\subsection{Training settings}
\label{app:training_settings}
In our experiments, we train the Qwen3-30B-A3B-Thinking-2507 model using our AceGRPO. 
We use MLE-Dojo~\cite{qiang2025mle} to construct our training dataset. We exclude the overlapping 68 MLE-bench tasks contained within MLE-Dojo. The final training dataset consists of 134 machine learning competition tasks.
The reinforcement learning training runs for 400 steps with a rollout batch size of 8 tasks per iteration, 8 response samples per task, and a global training batch size of 64.
We implement our training pipeline using the slime~\cite{slime_github} framework. The model was trained for two days on a cluster consisting of two nodes, each equipped with eight NVIDIA H200 GPUs. During the training phase, we utilized ten nodes, each containing eight NVIDIA RTX 4090 GPUs, to serve as execution environments for the experiments.

\subsection{Overview of MLE-Bench}
\label{app:mle_bench}

MLE-Bench~\cite{chan2024mle} is a benchmark designed to evaluate the machine learning engineering capabilities of AI agents. The dataset consists of 75 Kaggle competitions, curated from an initial pool of over 5,000, to assess real-world skills such as model training, dataset preparation, and experiment execution. These competitions span 15 diverse domains, including natural language processing, computer vision, and signal processing. To evaluate agents across varying levels of difficulty, tasks are categorized based on the estimated time required for an experienced human engineer to solve them: 30\% are Low complexity ($<$ 2 hours), 50\% are Medium (2--10 hours), and 20\% are High complexity ($>$ 10 hours).

MLE-Bench functions as an offline environment that replicates the rigorous conditions of actual Kaggle competitions. While original datasets are utilized whenever feasible, many competitions do not release private test sets. In such instances, new train-test splits are constructed from publicly available training data, typically reserving 10\% for testing while preserving the original data distribution. Agents are required to generate a standardized \texttt{submission.csv} file for all tasks, which is evaluated using a local grading system. To unify performance assessment across diverse raw metrics, the benchmark employs the Kaggle Medal system (Bronze, Silver, Gold). Thresholds are dynamically calculated based on the original human leaderboards to align with real-world competitive standards.

MLE-Bench employs a comprehensive set of evaluation metrics. The primary indicators are the rates of achieving Any Medal, stratified by task complexity, alongside the overall average medal rate. Furthermore, the evaluation reports the proportion of tasks with valid submissions, the percentage of tasks where the model outperforms the median human competitor, and the specific acquisition rates for Bronze, Silver, and Gold medals. The specific definitions for these metrics are as follows:

\begin{itemize}[leftmargin=1em,itemsep=0pt,parsep=0.2em,topsep=0.0em,partopsep=0.0em]
    \item \textbf{Valid Submission:} The participant uploads a submission file that strictly adheres to the formatting requirements specified by the competition.
    \item \textbf{Above Median:} A valid submission achieves a leaderboard score that surpasses the performance of at least 50\% of human participants.
    \item \textbf{Bronze Medal:} A valid submission achieves a leaderboard score that meets the official Kaggle threshold for a Bronze medal but does not reach the Silver threshold.
    \item \textbf{Silver Medal:} A valid submission achieves a leaderboard score that meets the official Kaggle threshold for a Silver medal but does not reach the Gold threshold.
    \item \textbf{Gold Medal:} A valid submission achieves a leaderboard score that meets or exceeds the official Kaggle threshold for a Gold medal.
    \item \textbf{Any Medal:} A valid submission achieves a leaderboard score that meets the requirements for at least a Bronze medal.
\end{itemize}

Given the prohibitive computational overhead and time constraints associated with a full evaluation on the complete MLE-Bench, the organizers of MLE-Bench have introduced MLE-Bench to facilitate more efficient benchmarking for the research community. This lightweight version comprises a curated subset of 22 Kaggle tasks selected from the original suite. Aside from the reduced task count, the evaluation protocols and metrics remain strictly consistent with the standard MLE-Bench.


\section{Hyperparameters}
  \label{app:hyperparameters}

This section summarizes the main hyperparameters used for AceGRPO training.
We report the algorithmic and optimization settings that materially affect the method, and omit infrastructure-only details such as sandbox retry limits, polling intervals, and logging options.

\subsection{Buffer and Adaptive Sampling}

\Cref{tab:buffer_sampling_params} lists the hyperparameters for the evolving data buffer and the adaptive sampling distribution in \Cref{sec:buffer,sec:sampling}.
AceGRPO uses a smooth temperature schedule and a recency factor rather than a hand-designed multi-stage curriculum.

\begin{table}[!t]
\centering
\small
\caption{Key hyperparameters for buffer construction and adaptive sampling.}
\label{tab:buffer_sampling_params}
\begin{tabular}{lccp{6.5cm}}
\toprule
\textbf{Parameter} & \textbf{Symbol} & \textbf{Value} & \textbf{Description} \\
\midrule
Initial potential & $\mathcal{P}_{\text{init}}$ & 0.1 & Default potential for newly inserted states before their first rollout \\
Std. deviation clip & $\delta_{\max}$ & 1.0 & Upper bound for the uncertainty term in the learnability potential \\
Initial temperature & $\tau_0$ & 1.0 & Initial temperature for potential-based sampling \\
Temperature floor & $\epsilon_\tau$ & 0.1 & Minimum temperature ratio in the annealing schedule \\
Saturation buffer size & $N_{\mathrm{sat}}$ & 128 & Buffer size at which the sampler reaches the temperature floor \\
Recency window & $\Delta$ & 10 & Number of rollout steps over which recently sampled states are down-weighted \\
Max buffer size & $N_{\max}$ & 128 & Maximum number of state groups maintained in the evolving buffer \\
Improvement reward weight & $\alpha$ & 0.3 & Weight of the relative-improvement term in \Cref{eq:shaped_reward} \\
\bottomrule
\end{tabular}
\end{table}

\subsection{GRPO Rollout Configuration}

\Cref{tab:grpo_params} lists the rollout-level settings used to construct each GRPO training batch.

\begin{table}[!t]
\centering
\small
\caption{GRPO rollout configuration.}
\label{tab:grpo_params}
\begin{tabular}{lcp{7.5cm}}
\toprule
\textbf{Parameter} & \textbf{Value} & \textbf{Description} \\
\midrule
Group size & $G=8$ & Number of responses sampled from each starting state \\
Rollout batch size & $B=8$ & Number of starting states sampled per rollout step \\
Global batch size & 64 & Number of generated samples used per policy update ($B\times G$) \\
Rollout temperature & 0.7 & Sampling temperature for policy generation \\
Maximum response length & 16,384 & Maximum number of generated tokens per response \\
Main training steps & 520 & Number of rollout/update iterations for the main AceGRPO run \\
\bottomrule
\end{tabular}
\end{table}

\subsection{Policy Optimization}

\Cref{tab:optimization_params} provides the policy optimization settings.

\begin{table}[h]
\centering
\small
\caption{Policy optimization hyperparameters.}
\label{tab:optimization_params}
\begin{tabular}{lcp{7.5cm}}
\toprule
\textbf{Parameter} & \textbf{Value} & \textbf{Description} \\
\midrule
Optimizer & Adam & Adaptive moment estimation optimizer \\
Learning rate & $1\times 10^{-6}$ & Constant learning rate for policy updates \\
Weight decay & 0.1 & L2 regularization coefficient \\
Adam $\beta_1$ & 0.9 & Exponential decay rate for the first moment estimate \\
Adam $\beta_2$ & 0.98 & Exponential decay rate for the second moment estimate \\
PPO clip range & $[0.2, 0.28]$ & Lower and upper clipping bounds used in the GRPO/PPO-style objective \\
KL loss coefficient & 0.005 & Coefficient for the KL penalty against the reference policy \\
Entropy coefficient & 0.0005 & Coefficient for entropy regularization \\
\bottomrule
\end{tabular}
\end{table}


\subsection{Hyperparameter Ablation}
\label{app:hyperparameter_ablation}

We only conduct hyperparameter ablation on the reward mixing coefficient $\alpha$ in \Cref{eq:shaped_reward}.
This is because $\alpha$ is the only hyperparameter that directly changes the learning objective: it controls the trade-off between optimizing absolute solution quality and encouraging relative improvement over the current state.
The sensitivity runs use the same Qwen3-30B-A3B-Thinking-2507 initialization, the same MLE-Dojo training set, the same two-node training layout, and the same AceGRPO configuration, changing only \texttt{reward.alpha\_improve}.
For efficiency, these runs are trained for 200 steps.

In contrast, the remaining hyperparameters in \Cref{tab:buffer_sampling_params,tab:grpo_params,tab:optimization_params} are not separately ablated.
Parameters such as $N_{\mathrm{sat}}$, $\Delta$, and $N_{\max}$ specify the operating scale of the buffer and sampler: $N_{\mathrm{sat}}$ determines when temperature annealing reaches its floor, $\Delta$ controls short-term recency suppression, and $N_{\max}$ bounds the training-side state pool.
Similarly, the group size, rollout batch size, and maximum response length determine the execution budget per update, while optimizer settings follow stable GRPO/PPO-style fine-tuning defaults.
Ablating these quantities would mostly mix algorithmic effects with compute budget, wall-clock throughput, memory pressure, or buffer-capacity constraints, rather than testing a distinct methodological claim.
We therefore use component ablations in the main text to evaluate the data buffer and adaptive sampler, and reserve hyperparameter sensitivity for $\alpha$.

\Cref{tab:alpha_ablation} summarizes the 200-step sensitivity study.
With $\alpha=0$, the reward reduces to the absolute HumanRank score and the Any Medal rate drops below the base model, suggesting that absolute quality alone provides insufficient pressure for iterative refinement.
Increasing the weight to $\alpha=0.6$ recovers performance and slightly improves over the base model, but still underperforms the default $\alpha=0.3$ setting at the same 200-step checkpoint.
This indicates that AceGRPO benefits from relative-improvement feedback, but overly emphasizing improvement over the parent state can make the reward more sensitive to local baselines and noisy small gains.
We therefore use $\alpha=0.3$ as the default setting, which provides the best trade-off in our runs.

\begin{table}[!t]
\centering
\small
\caption{Hyperparameter ablation for the reward mixing coefficient $\alpha$ on MLE-Bench. $\Delta$ denotes the Any Medal change relative to the base Qwen3-30B model. Dashes indicate metrics not included in this sensitivity summary.}
\label{tab:alpha_ablation}
\resizebox{\linewidth}{!}{
\begin{tabular}{lcccccc}
\toprule
\textbf{Model / Setting} & \textbf{Steps} & \textbf{Valid} & \textbf{Above Med.} & \textbf{HumanRank} & \textbf{Any Medal} & \textbf{$\Delta$} \\
\midrule
Qwen3-30B baseline & -- & 84.85 & 50.00 & 48.75 & 27.27 & -- \\
$\alpha=0.0$ & 200 & 92.42 & 46.97 & 46.52 & 19.70 & -7.57\%  \\
$\alpha=0.6$ & 200 & 93.18 & 50.00 & 51.69 & 29.55 & +2.28\% \\
$\alpha=0.3$ (AceGRPO) & \textbf{200} & \textbf{96.97} & \textbf{56.06} & \textbf{59.37} & \textbf{37.30} & \textbf{+10.03\%} \\
\bottomrule
\end{tabular}
}
\end{table}

\section{Limitation}
\label{app:limitation}
Due to the high execution cost of long-horizon MLE environments, our experiments are conducted under a constrained training budget and limited rollout scale. Larger-scale training and longer curriculum evolution may further improve performance, but were not explored in this work.

\newpage


\section{Prompts}
\label{app:prompts}

\begin{promptbox}[Prompts for drafting an initial code]
\label{app:prompts_draft}
You are a Kaggle grandmaster attending a competition. 
In order to win this competition, you need to come up with an excellent and creative plan for a solution and then implement this solution in Python. We will now provide a description of the task.\\

\# Task description

\{task\_description\}\\

\# Instructions

\#\# Response format

Your response should be a brief outline/sketch of your proposed solution in natural language (3-5 sentences), followed by a single markdown code block (wrapped in ```) which implements this solution and prints out the evaluation metric. There should be no additional headings or text in your response. Just natural language text followed by a newline and then the markdown code block.\\

\#\# Solution sketch guideline

- The solution sketch should be 3-5 sentences.

- Propose an evaluation metric that is reasonable for this task.

- Don't suggest to do EDA.

- The data is already prepared and available in the `./input` directory. There is no need to unzip any files.\\

\#\# Implementation guideline

- The code must not only implement the proposed solution but also **print the evaluation metric computed on a hold-out validation set**. **Without this metric, the solution cannot be evaluated, rendering the entire code invalid.**,

- **AND MOST IMPORTANTLY SAVE PREDICTIONS ON THE PROVIDED UNLABELED TEST DATA IN A `submission.csv` FILE IN THE ./submission/ DIRECTORY.**
- The code should be a single-file python program that is self-contained and can be executed as-is.

- No parts of the code should be skipped, don't terminate the before finishing the script.

- Your response should only contain a single code block.

- All the provided input data is stored in "./input" directory.

- **You MUST submit predictions on the provided unlabeled test data in a `submission.csv` file** file in the "./working" directory as described in the task description** This is extremely important since this file is used for grading/evaluation. DO NOT FORGET THE submission.csv file!

- You can also use the "./working" directory to store any temporary files that your code needs to create.

- REMEMBER THE ./submission/submission.csv FILE!!!!! The correct directory is important too.

- If you use `DataLoader`, you need to increase the parameter `num\_workers` to speed up the training process.\\

\#\# Installed Packages

Your solution can use any relevant machine learning packages such as: `pandas`, `statsmodels`, `torch-geometric`, `bayesian-optimization`, `torch`, `xgboost`, `spacy`, `timm`, `scikit-learn`, `transformers`, `nltk`, `lightGBM`, `numpy`, `torchvision`. Feel free to use any other packages too (all packages are already installed!). For neural networks we suggest using PyTorch rather than TensorFlow.\\

\# Data preview

\{data\_preview\}
\end{promptbox}

\begin{promptbox}[Prompts for debugging]
\label{app:prompts_debugging}
You are a Kaggle grandmaster attending a competition. Your previous solution had a bug and/or did not produce a submission.csv, or the generated submission.csv was in an incorrect format,so based on the information below, you should revise it in order to fix this. Your response should be an implementation outline in natural language, followed by a single markdown code block which implements the bugfix/solution.\\

\# Task description

\{task\_description\}\\

\# Instructions

\#\# Response format

Your response should be a brief outline/sketch of your proposed solution in natural language (3-5 sentences), followed by a single markdown code block (wrapped in ```) which implements this solution and prints out the evaluation metric. There should be no additional headings or text in your response. Just natural language text followed by a newline and then the markdown code block.\\

\#\# Bugfix improvement sketch guideline

- - You should write a brief natural language description (3-5 sentences) of how the issue in the previous implementation can be fixed.

- - Don't suggest to do EDA.

- - You should keep the core method of the machine learning code same. Do not change the machine learning method and just fix the code.

- - If the code failed because of missing library, try to avoid using the missing library. Do not try to install the missing library.

- - All packages have been installed. You are not allowed to install anything with pip or conda. If something is missing, try another way instead of installing a package.\\

\#\# Implementation guideline

- The code must not only implement the proposed solution but also **print the evaluation metric computed on a hold-out validation set**. **Without this metric, the solution cannot be evaluated, rendering the entire code invalid.**,

- **AND MOST IMPORTANTLY SAVE PREDICTIONS ON THE PROVIDED UNLABELED TEST DATA IN A `submission.csv` FILE IN THE ./submission/ DIRECTORY.**

- The code should be a single-file python program that is self-contained and can be executed as-is.

- No parts of the code should be skipped, don't terminate the before finishing the script.

- Your response should only contain a single code block.

- All the provided input data is stored in "./input" directory.

- **You MUST submit predictions on the provided unlabeled test data in a `submission.csv` file** file in the "./working" directory as described in the task description** This is extremely important since this file is used for grading/evaluation. DO NOT FORGET THE submission.csv file!

- You can also use the "./working" directory to store any temporary files that your code needs to create.

- REMEMBER THE ./submission/submission.csv FILE!!!!! The correct directory is important too.

- If you use `DataLoader`, you need to increase the parameter `num\_workers` to speed up the training process.\\

\# Data preview

\{data\_preview\}\\

\# Previous (buggy) implementation

\{buggy\_code\}\\

\# Execution output

\{terminal\_output\}
\end{promptbox}

\begin{promptbox}[Prompts for improving]
\label{app:prompts_improving}
You are a Kaggle grandmaster attending a competition. You are provided with previous memory including previously developed solutions and an creative idea. You need to implement this idea on top of (or building upon) the previously developed solution and memory.\\

\# Task description

Here is the original kaggle task description.

\{task\_description\}\\

\# Instructions

Here is the instruction about response format and implementation.

\#\# Response format

Your response should be a brief outline/sketch of the solution in natural language (3-5 sentences), followed by a single markdown code block (wrapped in ```) which implements this solution and prints out the evaluation metric. There should be no additional headings or text in your response. Just natural language text followed by a newline and then the markdown code block.\\

\#\# Solution improvement sketch guideline

- - The solution sketch should be a brief natural language description of how you improved the previous solution.

- - The solution sketch should be 3-5 sentences.

- - Don't do EDA.

- - All packages have been installed. You are not allowed to install anything with pip or conda. If something is missing, try another way instead of installing a package.\\

\#\# Implementation guideline

- The code must not only implement the creative idea but also **print the evaluation metric computed on a hold-out validation set**. **Without this metric, the solution cannot be evaluated, rendering the entire code invalid.**,

- **AND MOST IMPORTANTLY SAVE PREDICTIONS ON THE PROVIDED UNLABELED TEST DATA IN A `submission.csv` FILE IN THE ./submission/ DIRECTORY.**

- The code should be a single-file python program that is self-contained and can be executed as-is.

- No parts of the code should be skipped, don't terminate the before finishing the script.

- Your response should only contain a single code block.

- All the provided input data is stored in "./input" directory.

- **You MUST submit predictions on the provided unlabeled test data in a `submission.csv` file** file in the "./working" directory as described in the task description** This is extremely important since this file is used for grading/evaluation. DO NOT FORGET THE submission.csv file!

- You can also use the "./working" directory to store any temporary files that your code needs to create.

- REMEMBER THE ./submission/submission.csv FILE!!!!! The correct directory is important too.

- If you use `DataLoader`, you need to increase the parameter `num\_workers` to speed up the training process.\\

\# Data preview
\{data\_preview\}
\end{promptbox}



\end{document}